# Engineering Education in the Age of Autonomous Machines

Shaoshan Liu[1]  *Senior Member IEEE*
Jean-Luc Gaudiot[2]  *Fellow IEEE*
Hironori Kasahara[3]  *Fellow IEEE*

[1]PerceptIn Inc, U.S.A
[2]University of California, Irvine, U.S.A.
[3]Waseda University, Japan

Abstract:

*In the past few years, we have observed a huge supply-demand gap for autonomous driving engineers. The core problem is that autonomous driving is not one single technology but rather a complex system integrating many technologies, and no one single academic department can provide comprehensive education in this field.  We advocate to create a cross-disciplinary program to expose students with technical background in computer science, computer engineering, electrical engineering, as well as mechanical engineering. On top of the cross-disciplinary technical foundation, a capstone project that provides students with hands-on experiences of working with a real autonomous vehicle is required to consolidate the technical foundation.*

## Introduction

We have entered the age of autonomous machines, as increasing numbers of mobile robots, drones, and autonomous vehicles become part of our daily life [1]. While these autonomous machines greatly increase our productivity and efficiency, the industry of autonomous machines is facing an imminent problem of talent shortage.

Take autonomous driving for example, in the past few years, we have observed a huge supply-demand gap for autonomous driving engineers [2].  On the one hand, companies face great difficulties in hiring good autonomous driving engineers; on the other hand, the existing university engineering education does not prepare students with sufficient technical background to take on these opportunities. Instead, many students seek help from professional training agencies or online courses and many companies have to spend significant of time and efforts to train their incoming engineers [3].

The core problem is that autonomous driving is not one single technology but rather a complex system integrating many technologies, and no one single department can provide comprehensive education in this field [4]. To bridge this supply-demand gap, we advocate to create a cross-disciplinary program to expose students with technical background in computer science, computer engineering, electrical engineering, as well as mechanical engineering. On top of the cross-disciplinary technical foundation, a capstone project that provides students with hands-on experiences of working with a real autonomous vehicle is required to consolidate the technical foundation.

## The Rise of Autonomous Machines

It has been estimated that autonomous machine will gradually become a reality to the point of being ubiquitous by 2040. However, enormous research and development efforts must still be undertaken [1]. In the past few years, there has been a misconception that autonomous machines is a subfield of artificial intelligence and thus that autonomous machine engineers are all artificial intelligence engineers. Although artificial intelligence is an important piece of autonomous machines, autonomous machines also require complex systems engineering that consists of the many divisions of R&D across multiple academic departments [5]:

- Sensing: the main purpose of the sensing subsystem is to extract meaningful information from raw sensor data, including LiDAR, camera, radar, GPS, sonar, *etc,* which involves signal processing and optics. This is usually covered by the curriculum of the Electrical Engineering Department.

- Perception and Planning: the main purpose of the perception subsystem is for the autonomous machine to understand its surrounding environment, which involves traditional machine learning, deep learning, and other computer vision techniques. The planning subsystem is for vehicles to identify an optimal route and optimal trajectories, which involves optimization algorithms and even deep learning approaches. Both perception and planning are covered by the curriculum of the Computer Science Department.

- Client and Cloud Computing: the client computing subsystem consists of operating systems, computer architecture, and even VLSI design. The key challenge is to integrate various algorithms to meet real-time, reliability, safety, and energy consumption requirements. The cloud computing subsystem provides offline computing and storage capabilities to support testing new algorithms, generating high-definition maps, and training deep-learning models. It often requires knowledge in distributed computing, graphics, and big data. Both client computing and cloud computing are covered by the curriculum of the Computer Engineering Department.

- Mechanical Control: the main purpose of the mechanical control subsystem is to convert the planned trajectories into actual control commands*,* which involves linear systems, transfer functions, stability and feedback. This is covered by the curriculum of the Mechanical Engineering Department.

Hence, we conclude that with traditional academic department partitioning, no single academic department can provide comprehensive education for autonomous machine engineers. Instead, a cross-disciplinary approach that consists of at least four pillars, including electrical engineering, computer science, computer engineering, and mechanical engineering must be designed.

## The Supply-Demand Gap for Autonomous Machine Engineers

With traditional academic department organization, autonomous machine companies find it difficult to hire qualified fresh graduates from universities. Hence, in the past few years, industry has been plagued with a tremendous supply-demand gap for autonomous machine engineers. Indeed, according to [2], the rise of self-driving cars will create more than 100,000 U.S. jobs in the next decade, while the demand could be up to six times more than the expected number of graduates in this field.

On the supply side, many students actually have a high degree of interest in autonomous driving technologies but find that their home universities do not provide proper education to prepare them to enter this field. To address this problem, these students often have to seek help from professional training agencies or online training platforms such as the one described in [7]. Although these autonomous driving classes provide great introductory material to the subject, these classes usually focus on only one or two technologies, often deep learning related. Also, with these online classes, it is hard for the students to interact with each other or with the instructor. It is thus difficult to get useful feedback to improve the learning process. In addition, with these online classes, it is challenging for the students to learn how to integrate different modules into a working system. This disconnection between individual technology modules and system integration creates a very high entry barrier, and often "scares" interested students from this exciting field.

On the demand side, autonomous machine companies have to develop internal training programs to prepare incoming engineers for the job. For instance, to deal with the shortage of talent supply, in the past few years an autonomous driving company has established a two-week intensive training program to train incoming engineers [3, 6]. Most of the company's incoming engineers possess an embedded systems or general software engineering background. In the first week of the training, the incoming engineers all start with a technology overview, followed by the client systems and the cloud platforms modules. Since the engineers come from embedded systems and general software engineering background, they are quite comfortable with these modules. Through system modules, the incoming engineers also learn about the characteristics of different workloads as well as how to integrate them on embedded and cloud systems. In the second week, based on the students' performance in the first week as well as their interest levels towards different technologies, they are assigned to dig deeper into a specific module, such as perception, localization, or decision making. At the end of the two-week training, the incoming engineers are assigned to perform integration experiments to design an actual autonomous vehicle from scratch. Although the training period is short, this methodology has successfully onboarded engineers with limited prior background in autonomous driving.

In retrospect, the incoming engineers already possessed strong engineering foundations, such as math and basic engineering skills, through their university education. Also, out of the four pillars, electrical engineering, computer science, computer engineering, and mechanical engineering, the incoming engineers already had in-depth training in one or two of the pillars, acquired through both university education and prior work experiences. Within the two-week training period, the autonomous driving company thus first exposes these engineers with a broad spectrum of all technical modules involved in autonomous driving, and a capstone project that requires the engineers to bring up a real autonomous vehicle from scratch really consolidates all the technical modules and allows the engineers to integrate all the modules into a working system, and hence to gain a deeper understanding of the whole autonomous driving system domain.

## Cross-Disciplinary Education Program on Autonomous Machines

As indicated in the previous section, the autonomous machine knowledge stack consists of three layers: the first layer is the foundation; it includes basic math and engineering skills taught in freshman year and sophomore year in most engineering schools, regardless of the specific academic department. The second layer consists of the pillars; it includes the advanced and specialized engineering classes taught in traditional junior and senior years. The third layer is the capstone project that enables the students to integrate all the technical modules into a working system. It is important to note that this layer is not available in most universities and best taught at autonomous machine companies.

The key observation here is that most of the technical foundation needed for autonomous machine education and pillars, except for the capstone project, is already being taught: these students share the same engineering foundation but the pillars are taught across different departments. Therefore, for universities to bridge the supply-demand gap of autonomous machine engineers, we advocate the creation of cross-disciplinary programs, probably a one-year graduate degree program or a minor degree in undergraduate education. For students with a technical background in electrical engineering, computer science, computer engineering, and mechanical engineering, one or two classes in each pillar would be taught in the context of creating an autonomous machine. This would serve to finalize their knowledge in this more specialized field. The core mission of these classes is not only to teach students the theories behind these fields, but more importantly to enable students to grasp the connections between these fields and eventually to think in a systematic way how autonomous machines work. The last piece of the program would be a capstone project where the students would design an autonomous machine from scratch, so that they can understand how the knowledge learned in classes can be applied in creating a real-world autonomous machine. We call this approach the Foundation-Pillars-Capstone.

From the past four years of experiences of onboarding engineers with limited background in autonomous machines, the Foundation-Pillars-Capstone approach has worked very well. However, to adopt this approach in a university environment, the key challenge is the design of the capstone project as universities often lack expertise in developing commercial-grade autonomous machines. Note that, the capstone project not only equips the students with the necessary knowledge to work on autonomous machines, but more importantly provides the students with the systematic insight of how different pillars interconnect with each other, this prepares the students for lifelong technical challenges beyond autonomous machines. Indeed, it is incumbent upon educators to recognize that universities are to "educate" the students and their thinking rather than "train" them and their skills for a possibly ephemeral purpose. This is why it remains paramount to build this education upon solid theoretical foundations which will permit the students to evolve in their technical careers and refocus their skills with technology changes.

## Policy Recommendations

To address the capstone project problem, cooperative education (co-op), a structured method combining classroom-based education with practical work experience, might be a suitable solution [8]. In this arrangement, universities can collaborate with industry partners to have students perform the capstone project at the autonomous machine company. Different autonomous machine companies could include mobile robots, drones, autonomous vehicles, *etc.* With the co-op approach, the students can choose the type of autonomous machines on which they want to perform the capstone project.

However, a downside with the co-op approach is that companies have to spend resources training the students with the potential of hiring them later. However, this may not be a strong enough incentive for the autonomous machine companies which are often early-stage companies with limited resources. To promote the collaboration between the industry and the academia in emerging fields like autonomous machines, we would need to seek government funding to provide incentives through appropriate grants programs, especially as the U.S. government has been playing an active in driving the U.S. innovations [9]. These education and training grants advance the domestic workforce skills, and consequently improve the domestic economy. For instance, one such program, the Strengthening Community Colleges Training Grants Program, provided by the U.S. Department of Labor, aims to build the capacity of community colleges to collaborate with employers and the public workforce development system to meet local and regional labor market demand for a skilled workforce [10].

If the government can provide incentives funds to promote industry-academia collaboration in advanced fields, such as autonomous machines, we can complete the close-looped education system that enable students to master the foundation, the pillars, and the capstone. This means that, immediately after training, students can more easily enter the workforce and contribute.

## Conclusions

Although we have great university systems around the globe, the reality is that a huge and widening supply-demand gap currently exists for autonomous machine engineers. In this article, we have identified a fundamental issue from both industry and academic perspectives, and we have proposed a solution to the widening supply-demand gap problem. We have further advocated for an increased government role by providing incentives funds to promote industry-academia collaboration in developing talents in advanced fields, such as autonomous machines.

## References:

1. Liu, S., Peng, J. and Gaudiot, J.L., 2017. Computer, drive my car!. Computer, (1), pp.8-8.
2. Self-driving cars will create 30,000 engineering jobs that the US can't fill, TechRepublic, https://www.techrepublic.com/article/self-driving-cars-will-create-30000-engineering-jobs-that-the-us-cant-fill/, accessed 12/5/2020
3. Tang, J., Shaoshan, L., Pei, S., Zuckerman, S., Chen, L., Shi, W. and Gaudiot, J.L., 2018, July. Teaching autonomous driving using a modular and integrated approach. In 2018 IEEE 42nd Annual Computer Software and Applications Conference (COMPSAC) (Vol. 1, pp. 361-366). IEEE.
4. Liu, S., Li, L., Tang, J., Wu, S. and Gaudiot, J.L., 2020. Creating Autonomous Vehicle Systems. Synthesis Lectures on Computer Science, 8(2), pp.i-216.
5. Liu, S., 2020. Engineering Autonomous Vehicles and Robots: The DragonFly Modular-based Approach. John Wiley & Sons.
6. Liu, S. and Gaudiot, J.L., 2020. Autonomous vehicles lite self-driving technologies should start small, go slow. IEEE Spectrum, 57(3), pp.36-49.
7. Self-Driving Cars Specialization, Coursera, https://www.coursera.org/specializations/self-driving-cars, accessed 12/5/2020
8. Blair, B.F., Millea, M. and Hammer, J., 2004. The impact of cooperative education on academic performance and compensation of engineering majors. Journal of Engineering Education, 93(4), pp.333-338.
9. Liu, S., 2020. DARPA: A Global Innovation Differentiator. IEEE Engineering Management Review.
10. Strengthening Community Colleges Training Grants Program, U.S. Department of Labor, https://www.dol.gov/agencies/eta/skills-training-grants/scc, accessed 12/5/2020